\documentclass[sigconf]{acmart}

\usepackage{graphicx}
\usepackage{listings}
\usepackage{amsmath}  
\usepackage{makecell}
\usepackage{booktabs}
\usepackage{multirow}
\usepackage{adjustbox}

\usepackage[most]{tcolorbox}
\usepackage{caption}
\usepackage{lipsum} 

\AtBeginDocument{%
  }

\setcopyright{acmlicensed}
\copyrightyear{2025}
\acmYear{2025}
\acmDOI{XXXXXXX.XXXXXXX}
\acmConference[SIGIR2025 LiveRAG Challenge]{SIGIR2025 LiveRAG Challenge}{February 2025}{TBD}
\acmISBN{9781450312345} 

\begin{document}
\title{DoTA-RAG: Dynamic of Thought Aggregation RAG}

\author{Saksorn Ruangtanusak}
\affiliation{%
  \institution{SCBX}
  \city{Bangkok}
  \country{Thailand}}
\email{saksorn.r@scbx.com}

\author{Natthapath Rungseesiripak}
\affiliation{%
  \institution{SCBX}
  \city{Bangkok}
  \country{Thailand}}
\email{natthapath.r@scbx.com}

\author{Peerawat Rojratchadakorn}
\affiliation{%
  \institution{SCBX}
  \city{Bangkok}
  \country{Thailand}}
\email{peerawat.r@scbx.com}

\author{Monthol Charattrakool}
\affiliation{%
  \institution{SCBX}
  \city{Bangkok}
  \country{Thailand}}
\email{monthol.c@scbx.com}

\author{Natapong Nitarach}
\affiliation{%
  \institution{SCB 10X}
  \city{Bangkok}
  \country{Thailand}}
\email{natapong@scb10x.com}

\renewcommand{\shortauthors}{Ruangtanusak et al.}

\begin{abstract}
In this paper, we introduce \textbf{DoTA-RAG} (\emph{Dynamic-of-Thought Aggregation RAG}), a Retrieval-Augmented Generation system optimized for high throughput, large-scale web knowledge indexes. Traditional RAG pipelines often suffer from high latency and limited accuracy over massive, diverse datasets. DoTA-RAG addresses these challenges with a three-stage pipeline: query rewriting, dynamic routing to specialized sub-indexes, and multi-stage retrieval and ranking. We further enhance retrieval by evaluating and selecting a superior embedding model, re-embedding the large FineWeb-10BT corpus. Moreover, we create a diverse Q\&A dataset of 500 questions generated via the DataMorgana setup across a broad range of WebOrganizer topics and formats. DoTA-RAG improves the answer correctness score from 0.752 (baseline, using LiveRAG pre-built vector store) to 1.478 while maintaining low latency and achieves a 0.929 correctness score on the Live Challenge Day. These results highlight DoTA-RAG’s potential for practical deployment in domains requiring fast, reliable access to large and evolving knowledge sources.
\end{abstract}


\maketitle

\begin{figure}[t]
\centering
\includegraphics[width=\columnwidth]{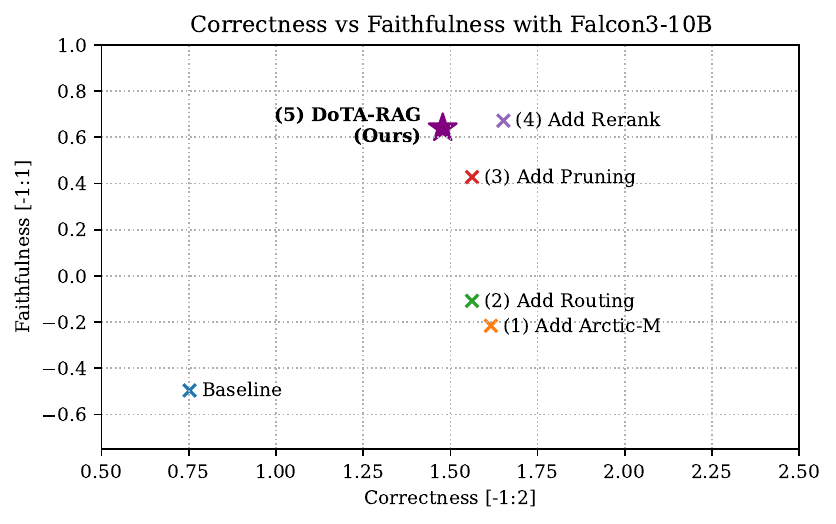}
\vspace{-7mm}
\caption{RAG Correctness and Faithfulness
in the internal test set for different retrieval-augmented generation approaches. We use the Falcon3-10B-Instruct as the base LLM}
\label{fig:chart_corr_faith}
\vspace{-7mm} 
\end{figure}

\section{INTRODUCTION AND RELATED WORK}

\begin{figure*}[t]
\centering
\includegraphics[scale=0.7]{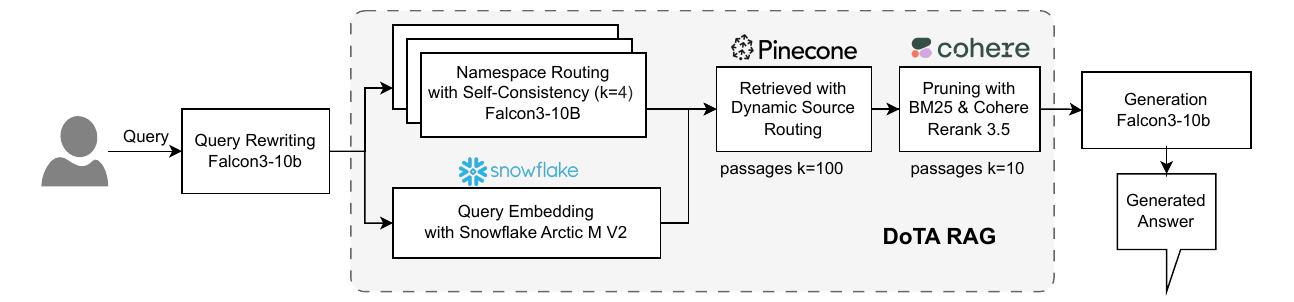}
\vspace{-3mm}
\caption{Diagram illustrating the components and workflow of DoTA-RAG.}
\vspace{-3mm}
\label{fig:dotarag_diagram}
\end{figure*}
Large language models (LLMs) have achieved strong results across NLP tasks, yet they often struggle with up-to-date or domain-specific knowledge, leading to hallucinations \cite{gao2024survey}. Retrieval-Augmented Generation (RAG) mitigates this by allowing LLMs to access external documents: a retriever fetches relevant text, which the LLM then conditions on when generating answers \cite{lewis2020rag}. Since its introduction by Lewis et al., RAG has become central to knowledge-intensive applications.

To improve retrieval robustness, \emph{query rewriting} has emerged as a key technique \cite{gao2024survey}, reformulating ambiguous or underspecified queries via prompt-engineered LLMs. For instance, HyDE prompts an LLM to generate a hypothetical answer document, enhancing retrieval by focusing on answer–answer similarity \cite{gao2022hyde}.

In the LiveRAG competition, we leverage the FineWeb-10BT corpus \cite{fineweb2024}. To handle its diverse sources, \emph{routing techniques} are employed to select optimal retrieval paths based on query semantics or metadata \cite{li2023semantic_route, modi2023autometa}. \emph{WebOrganizer} \cite{weborganizer2024} enhances this by categorizing documents by topic and format, enabling precise sub-corpus selection.

To reduce retrieved document count, we use \emph{rerankers}—cross-encoder models that reorder results based on deeper query-document interactions. Inspired by ColBERT \cite{khattab2020colbert}, production systems like Cohere’s Rerank 3.5 \cite{cohere2024rerank} improve relevance and filter noise.

Benchmarking RAG systems remains difficult due to the lack of realistic, diverse test sets. \emph{DataMorgana} \cite{datamorgana2023} addresses this with configurable tools for generating synthetic Q\&A datasets that reflect varied user intents.

The SIGIR 2025 LiveRAG Challenge mandates a fixed 15M-document corpus (FineWeb-10BT) and a standardized LLM (Falcon-3-10B-Instruct \cite{Falcon3}), posing significant challenges for scalable and accurate retrieval.

We propose DoTA-RAG (Dynamic of Thought Aggregation RAG) to address these issues. It integrates \textit{dynamic routing} and \textit{hybrid retrieval} to enhance retrieval efficiency and accuracy at scale.

In summary, our pipeline is built around two key components:
\begin{description}
\item[\textbf{Dynamic Routing:}] A framework that optimizes the retrieval path for large-scale corpora, reducing latency and enabling aggregation from diverse sources.
\item[\textbf{Hybrid Retrieval:}] A multi-stage strategy combining dense expansion, sparse filtering, and reranking for high-relevance document selection.
\end{description}

\section{DoTA-RAG}
\label{subsec:pipeline}

This section presents the production \textbf{DoTA-RAG} stack deployed for the SIGIR 2025 LiveRAG Challenge. We provide a detailed breakdown of each component in the end-to-end RAG pipeline as shown in Figure ~\ref{fig:dotarag_diagram}, highlighting the enhancements implemented to raise performance and reduce inference latency. In particular, we demonstrate how the system efficiently retrieves relevant passages from a knowledge base containing 15 million documents, consistently delivering responses in under one minute per query.

\subsection{Query Rewriting (Stage 1)}
\label{subsec:query_rewriting}
\paragraph{Low-Temp Rewrite Query.}  

Initially, we excluded a query rewriting stage, since methods like HyDE~\cite{gao-etal-2023-precise} and Chain-of-Draft~\cite{xu2025chaindraftthinkingfaster}, along with crafted prompts, were heavily tested but decreased Correctness and Faithfulness. They frequently did not generalize well or address internal retrieval challenges.

On Live Challenge Day, we discovered new failure cases involving user queries with either highly specialized terms or significant misspellings. This led to few or irrelevant results (e.g., \emph{"wut iz rajun cajun crawfsh festivl"}, \emph{"wut r sum side affects of nicotine gum"}), prompting us to re-evaluate query rewriting to tackle these issues. Appendix ~\ref{appendix:prompt} details the query rewriting prompt.

\vspace{-1.5mm}
\subsection{Dynamic Namespace Routing (Stage 2)}
\paragraph{Ensemble-Based Query Routing}
To route queries efficiently across topical domains, we maintain a separate Pinecone namespace for each domain (see Appendix~\ref{appendix:fineweb_characteristic}). For every incoming question, we generate four independent classifications using Falcon3-10B-Instruct with self-consistency voting~\cite{wang2023selfconsistencyimproveschainthought}. Each returns one or more namespace predictions, and we tally which namespaces appear most often. We then query the top two namespaces in parallel, shrinking the average search space by 92\% and reducing dense-retrieval latency from 100.84 s to 19.01 s per question. The prompt used for namespace routing can be found in Appendix~\ref{appendix:prompt}.

\vspace{-1.5mm}
\subsection{Hybrid Retrieval (Stage 3)}
\begin{enumerate}
\item \textbf{Dense search.} Embed the refined query with Snowflake Arctic-embed-m-v2.0 \cite{snowflake2024arctic} and fetch k=100 candidates using cosine similarity.
\item \textbf{BM25 pruning.} Compute lexical scores on-the-fly; retain the top 20 passages.
\item \textbf{Cohere's Rerank 3.5} Cross-encode and select the 10 highest-scoring passages.
\end{enumerate}

\vspace{-5mm}
\subsection{Context Aggregation (Stage 4)}
\label{subsec:context}
Concatenate the text of the top-10 passages, separated by blank lines.  
If the concatenation exceeds 8k tokens, truncate passages proportionally.

\vspace{-1.5mm}
\subsection{Answer Generation (Stage 5)}
\label{subsec:answer}
For answer generation, we use the prompt shown in Appendix~\ref{appendix:prompt}, which is prepended with the aggregated context and the rewritten input query. The answer is then generated by Falcon3-10B-Instruct \cite{Falcon3}.

\vspace{-1.5mm}
\paragraph{Section Summary.}
DoTA-RAG is a RAG pipeline developed for the SIGIR 2025 LiveRAG Challenge, designed to deliver high-quality answers with significantly reduced inference latency over a 15M-document knowledge base. The system features five key stages: (1) Query Rewriting, which enhances retrieval for noisy or misspelled queries; (2) Dynamic Namespace Routing, using an ensemble classifier to dynamically select relevant sub-indexes and shrink the search space; (3) Hybrid Retrieval, combining dense search, BM25 pruning, and cross-encoder re-ranking; (4) Context Aggregation, which compiles and trims top-ranked passages; and (5) Answer Generation, where Falcon3-10B-Instruct \cite{Falcon3} produces answers grounded in the retrieved context.

\vspace{-1.5mm}
\section{EXPERIMENTAL SETUP AND EVALUATION}
\label{sec:experiments}
This section details how we built our evaluation benchmark and which systems we compare against.

\vspace{-1.5mm}
\subsection{Internal test set Construction}

To create our internal test set, we utilized the \textbf{DataMorgana API} \cite{datamorgana2023} to generate an initial collection of 1,000 Q\&A pairs. To ensure question diversity, we adopted DataMorgana's categorization framework (user types, phrasing styles, premises, and linguistic variations) and enhanced it with our own question-formulation taxonomy (see Table \ref{tab:question_formulation}). This new categorization introduces carefully defined question types with varying complexity levels, including questions requiring either single or multiple document sources for resolution. 

For example, ``Temporal-evolution'' questions require the retrieval system to distinguish between documents discussing the same entity across different time periods, while challenging the generation module to synthesize information about chronological changes from these temporally distinct sources. While "Verification" questions test the system's ability to evaluate mixed assertions containing both facts and falsehoods. This category directly challenges the generation module to distinguish truth from misinformation, correct user misconceptions, and maintain factual integrity even when prompted with partially incorrect premises. Appendix \ref{appendix:question_formulation} provides descriptions and examples of the newly defined categories.

\begin{table}[H]
  \centering\small
  \caption{The question‐formulation categorization.}
  \vspace{-3mm}
  \label{tab:question_formulation}
  \begin{tabular}{@{} l r @{}}
    \toprule
    \textbf{Category}       & \textbf{Required Document(s)} \\ 
    \midrule
    Multi‐aspect            & Two documents                  \\ 
    Comparison              & Two documents                  \\ 
    Temporal‐evolution      & Two documents                  \\ 
    Problem‐solution        & Two documents                  \\ 
    Procedural              & Single document                \\ 
    Causal                  & Single document                \\ 
    Quantitative            & Single document                \\ 
    Verification            & Single document                \\ 
    \bottomrule
  \end{tabular}
\end{table}
\vspace{-2mm}

Each Q\&A pair's document was auto-tagged using \textbf{WebOrganizer}'s \href{https://huggingface.co/WebOrganizer/TopicClassifier}{TopicClassifier} (24 topics) and \href{https://huggingface.co/WebOrganizer/FormatClassifier}{FormatClassifier} (24 formats) \cite{weborganizer2024}. This tagging process was instrumental in creating \textbf{MorganaMultiDocQA}—our 500-question benchmark. The classification ensured diverse document topics and formats throughout the test set, which is critical for robust evaluation of retrieval-augmented models. This diversity prevents benchmark bias toward specific domains or document structures, thereby offering a more comprehensive assessment of model generalization capabilities across the heterogeneous content types found on the internet. We then applied stratified sampling to maintain proportional representation across all topic-format combinations, preserving the natural distribution patterns found in broader information ecosystems:%
\begin{equation}
  n_{c} \;=\;
  \Bigl\lceil
    \frac{N_{c}}{\sum_{c' \in \mathcal{C}} N_{c'}} \times 500
  \Bigr\rceil ,
\end{equation}
where $N_{c}$ is the number of candidates in category $c$ and $n_{c}$ the final sample size.  This guarantees balanced coverage across the $|\mathcal{C}| = 24 \times 24$ strata.

\subsection{Embedding Models (MTEB Leaderboard)}

To select a suitable dense retriever for our pipeline, we considered embedding model performance using the MTEB English retrieval tasks leaderboard \cite{muennighoff2023mteb}. Among models with fewer than 1 billion parameters, Snowflake's Arctic v2.0 embedding models demonstrated state-of-the-art results \cite{snowflake2024arctic}, achieving mean scores of 58.56 (large) and 58.41 (medium), significantly outperforming our initial model, E5-base-v2 \cite{wang2024e5}, which scored 49.67. Based on these findings, we re-embedded and re-indexed the FineWeb-10BT corpus \cite{fineweb2024} using Arctic-embed-m-v2.0. On our internal test set, this change led to an improvement in retrieval quality, with Recall@10 increasing from 0.469 to 0.518.



\subsection{Evaluation Metrics}

We assess system performance using two main metrics: \textit{Correctness} and \textit{Faithfulness}, each evaluated on a continuous scale.

\begin{itemize}
    \item \textbf{Correctness} (\textbf{-1} to \textbf{2}): Assesses the relevance and coverage of the generated answer. A score of \textbf{-1} indicates an incorrect answer, while \textbf{2} represents a fully correct and relevant response with no extraneous information.
    \item \textbf{Faithfulness} (\textbf{-1} to \textbf{1}): Measures whether the answer is grounded in the retrieved passages. A score of \textbf{-1} indicates no grounding at all, whereas \textbf{1} means the entire answer is fully supported by retrieved content.
\end{itemize}

\subsection{LLM-as-a-Judge Evaluation}
We graded answers for correctness and faithfulness using two automatic judges. While Claude 3.5 Sonnet \cite{anthropic2024sonnet} served as a strong judge, we found Falcon3-10B-Instruct \cite{Falcon3} to be a faster and cheaper alternative with comparable evaluation quality. During our experiments for the competition, we frequently used Falcon3-10B-Instruct to lower evaluation costs and accelerate the process. On the MorganaMultiDocQA test set, Falcon3-10B-Instruct yielded slightly higher scores on both metrics (see Table \ref{tab:judge}).

\begin{table}[h]
\caption{Judge agreement on the test set.}
\label{tab:judge}
\centering\small
\begin{tabular}{lrr}
\toprule
\textbf{Metric} 
& \makecell{\textbf{Claude 3.5}\\\textbf{Sonnet}~\cite{anthropic2024sonnet}} 
& \makecell{\textbf{Falcon3-10B}\\\textbf{Instruct}~\cite{Falcon3}} \\
\midrule
Correctness [-1:2]   & 1.382 & 1.430 \\
Faithfulness [-1:1]  & 0.520 & 0.580 \\
\bottomrule
\end{tabular}
\end{table}

\section{RESULTS AND ANALYSIS}
\label{sec:result_and_analysis}
This section presents our RAG pipeline evaluation across internal and live benchmarks. We report the incremental effects of each ablation step on correctness and faithfulness, and analyze final performance on the LiveRAG Live Challenge Day leaderboard.

\subsection{Internal test set Result}

Our ablation path incrementally augments the baseline pipeline (E5-base-v2 embeddings \cite{wang2024e5} + Falcon3-10B-Instruct \cite{Falcon3} generation) with various enhancements to isolate their individual and combined effects on performance:

\begin{description}
\item[\textsc{Baseline}] Intfloat's \texttt{e5-base-v2} → 10 retrieved passages → Falcon3‑10B-Instruct generation.
\item[\textsc{+Arctic‑M}] Swap in Snowflake's \texttt{arctic‑embed‑m‑v2.0} \cite{snowflake2024arctic}.
\item[\textsc{+Routing}] Dynamic routing for namespace selection.
\item[\textsc{+Pruning}] Retrieve 100 passages, then reduce to 10 passages using BM25.
\item[\textsc{+Rerank}] Prune to 20 passages, then use Cohere's Rerank 3.5 \cite{cohere2024rerank} to select the top 10.
\item[\textsc{+Rewrite}] Rewrite query to make the RAG pipeline more robust to the live test set – this is \textbf{DoTA‑RAG}.
\end{description}

\begin{table*}[htbp]
  \centering
  \normalsize
  \setlength\tabcolsep{12pt}
  \renewcommand\arraystretch{1.2}
  \caption{Pipeline's performance and inference time on internal test set using Claude 3.5 Sonnet \cite{anthropic2024sonnet}. Metrics are shown for full text and truncated inputs.}
  \vspace{-2mm}
  \label{tab:pipeline}
  \begin{tabular}{lccccc}
    \toprule
    \multirow{2}{*}{\textbf{Method}} 
    & \multicolumn{2}{c}{\textbf{Correctness [-1:2]}} 
    & \multicolumn{2}{c}{\textbf{Faithfulness [-1:1]}} 
    & \multirow{2}{*}{\textbf{Sec/Question}} \\
    \cmidrule(lr){2-3} \cmidrule(lr){4-5}
    & \textbf{All words} & \textbf{300-word cap} 
    & \textbf{All words} & \textbf{300-word cap} & \\
    \midrule
    Baseline         & 0.752 & 0.761 & -0.496 & -0.493 & -- \\
    + Arctic-M       & 1.616 & 1.626 & -0.216 & -0.225 & 100.84 \\
    + Routing        & 1.562 & 1.577 & -0.108 & -0.090 & 19.01 \\
    + Pruning        & 1.562 & 1.566 &  0.428 &  0.404 & 29.84 \\
    \textbf{+ Rerank}& \textbf{1.652} & \textbf{1.686} & \textbf{0.672} & \textbf{0.662} & \textbf{35.20} \\
    + Rewrite        & 1.478 & 1.484 &  0.640 &  0.620 & 35.63 \\
    \bottomrule
  \end{tabular}
\end{table*}

The results in Figure~\ref{fig:chart_corr_faith} and Table~\ref{tab:pipeline} illustrate the incremental improvements from each ablation step on the internal test\footnote{Note that inference time (Sec/Question) is omitted for the Baseline, as it uses a pre-built index that is not directly comparable to our FineWeb-based Pinecone index.} set. Replacing the baseline embeddings with \texttt{arctic-embed-m-v2.0}~\cite{snowflake2024arctic} (\textsc{+Arctic-M}) yields a substantial increase in correctness, though faithfulness remains negative. The addition of dynamic routing (\textsc{+Routing}) and BM25-based pruning (\textsc{+Pruning}) maintains high correctness and leads to a notable gain in faithfulness. Incorporating Cohere's Rerank 3.5~\cite{cohere2024rerank} (\textsc{+Rerank}) further improves both correctness, representing the strongest overall performance. Although adding rewriting decreases performance, we believe including it as part of the DoTA-RAG would better align our model with the test set in the LiveRAG Live Challenge Day, as discussed in Section~\ref{subsec:query_rewriting}.

\subsection{LiveRAG Live Challenge Day Performance}

On the official Live Challenge Day leaderboard, our final pipeline achieved a  
\textbf{correctness score of 0.929}, confirming that our retrieval–generation loop produced
high-quality answers.  
However, because the evaluation enforced a strict
\textit{300-word output cap}—a constraint we overlooked during tuning, our  
faithfulness score sank to just \textbf{0.043}. 

After the score was announced, it was surprising to see our faithfulness score so low. To understand what went wrong, we decided to conduct a deeper investigation. We re-ran the evaluation using Claude 3.5 Sonnet as the judge \cite{anthropic2024sonnet}, with a prompt tailored to the faithfulness criteria. We reassessed the 500 answers, both uncut and capped at 300 words. The faithfulness score dropped significantly from 0.702 to 0.336.

\section{LIMITATIONS AND FUTURE WORK}
\label{sec:futurework}
In future research endeavors, our goal is to explore strategies for multi-source routing that leverage graph-based knowledge bases, delving into the complexities of these systems. Furthermore, we are planning to investigate self-improvement methods applied after generating responses, aiming to refine this process. Another avenue of our research involves developing techniques for compacting context within windows larger than 8,000 tokens, as well as honing the procedures involved in reasoning retrieval, thereby enhancing overall performance.

\section{CONCLUSION}
\label{sec:conclusion}

We introduce \textbf{DoTA-RAG}, a live Retrieval-Augmented Generation pipeline that couples \emph{Dynamic‐of‐Thought Aggregation} with dynamic routing and hybrid retrieval.
By extracting on-the-fly metadata and steering each query to the most appropriate sub-index, DoTA-RAG reconciles the seemingly contradictory requirements of \emph{web-scale knowledge integration}, \emph{precision}, and \emph{low latency} that characterize the SIGIR 2025 LiveRAG Challenge.

Experiments on our 500-question \textsc{MorganaMultiDocQA} benchmark demonstrate that successive upgrades—from changing the dense retriever to Arctic-embed-m-v2.0 \cite{snowflake2024arctic} to incorporating query rewriting—increase correctness from 0.752 to 1.478 and faithfulness from -0.496 to 0.640, while maintaining a median end-to-end latency of 35.63 seconds per question. On the Live Challenge Day dataset, the same configuration achieves 0.929 correctness, validating the generalization of our RAG beyond in-house data.

Our ablation studies highlight two actionable insights for the Gen-IR community:

\begin{enumerate}
\item \textbf{Metadata-guided routing delivers outsized gains.} Even lightweight routing cues significantly reduce irrelevant results and cut retrieval latency by more than half compared to static top-$k$ search.
\item \textbf{Hybrid retrieval significantly enhances document quality.} By combining a fan-out dense retriever for broad semantic coverage and a sparse retriever for keyword-based filtering, the system improves the faithfulness score from -0.108 to 0.428 — a substantial gain in factual alignment.
\end{enumerate}

\section*{Acknowledgments}
We appreciate the SIGIR 2025 LiveRAG Challenge organizers for supplying the resources and tools needed for our participation and the evaluation set creation.

\newpage
\bibliographystyle{ACM-Reference-Format}
\bibliography{scbx_final_ref}

\newpage
\appendix

\section{Prompt}
\label{appendix:prompt}

Effective prompt engineering is critical for leveraging large language models (LLMs) in a variety of tasks, from rewriting text to classifying queries and generating informative responses. In this section, we outline carefully designed prompt templates that guide the behavior of LLM-powered assistants. These prompts are constructed to ensure clarity, precision, and relevance in model outputs, enhancing both usability and accuracy across diverse scenarios, as illustrated by the examples in Figure~\ref{fig:rewrite-prompt} (Query Rewriting), Figure~\ref{fig:routing-prompt} (Routing Namespace Classify), and Figure~\ref{fig:gen-prompt} (Generation Prompt).

\vspace{-3mm}
\begin{figure}[ht]
\centering
\begin{tcolorbox}[
  colback=white,
  colframe=black!75!black,
  title=Query Rewriting,
  fonttitle=\bfseries,
  fontupper=\ttfamily
]
You are a helpful assistant. 
Your task is to rewrite sentences by correcting typos 
and improving the wording to ensure they are 
written in clear, natural English. 
If a typo is intentional or acceptable as-is, 
leave it unchanged.
\end{tcolorbox}
\vspace{-3mm}
\caption{Query Rewriting}
\label{fig:rewrite-prompt}
\end{figure}

\vspace{-3mm}
\begin{figure}[ht]
\centering
\begin{tcolorbox}[
  colback=white,
  colframe=black!75!black,
  title=Routing Namespace Classify,
  fonttitle=\bfseries,
  fontupper=\ttfamily
]

Q: \{question\}

Available namespaces:
\{choices\_str\}

\texttt{Step1: Identify what the question is about.}
\texttt{Step2: Choose only the most relevant namespaces.}
\texttt{Step3: Return final result in //boxed\{\}.}
\end{tcolorbox}
\vspace{-3mm}
\caption{Routing Namespace Classify}
\label{fig:routing-prompt}
\end{figure}

\vspace{-3mm}
\begin{figure}[ht]
\centering
\begin{tcolorbox}[
  colback=white,
  colframe=black!75!black,
  title=Generation prompt,
  fonttitle=\bfseries,
  fontupper=\ttfamily
]
\texttt{You are a helpful assistant.} \\ \\
\texttt{Context:} \\
\texttt{<passages>} \\ \\
\texttt{Question:} \\
\texttt{<query>}
\end{tcolorbox}
\vspace{-3mm}
\caption{Generation Prompt}
\label{fig:gen-prompt}
\end{figure}

\begin{figure*}[t]
    \centering
    \includegraphics[width=\textwidth]{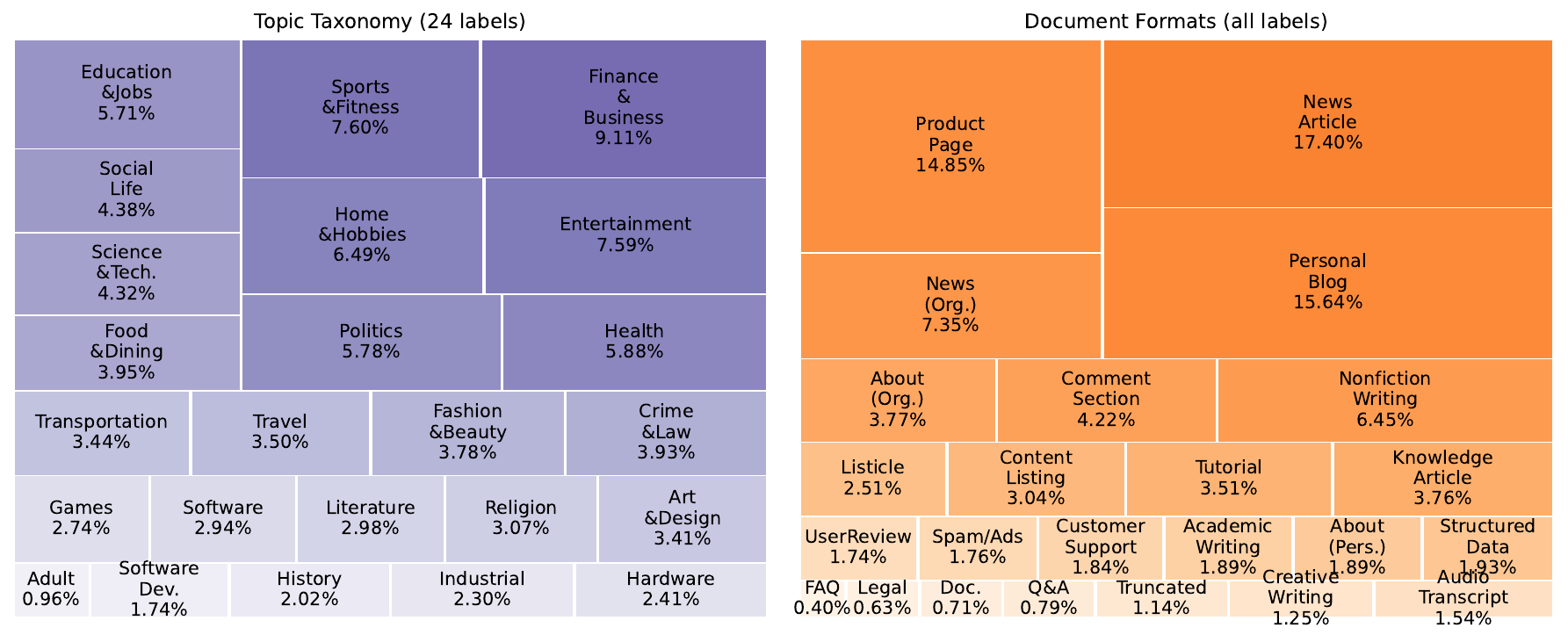}
    \vspace{-3mm}
    \caption{Distribution of topics (left) and document formats (right) in Fineweb-10BT, based on WebOrganizer classifiers. The area of each block reflects the number of documents per domain in the corpus.}
    \vspace{-3mm}
    \label{fig:dist_of_topics_format}
\end{figure*}

\section{Categorization and Description of Question Formulations with Examples}
\label{appendix:question_formulation}

\paragraph{Multi-aspect} A question about two different aspects of the same entity/concept. For example: ``What are the advantages of AI-powered diagnostics, and what are the associated risks of bias in medical decision-making?'', ``How do cryptocurrencies enable financial inclusion, and what are the security risks associated with them?''. The information required to answer the question needs to come from two documents; specifically, the first document must provide information about the first aspect, while the second must provide information about the second aspect.

\paragraph{Comparison} A comparison question that requires comparing two related concepts or entities. The comparison must be natural and reasonable, i.e., comparing two entities by a common attribute that is meaningful and relevant to both entities. For example: ``Who is older, Glenn Hughes or Ross Lynch?'', ``Are Pizhou and Jiujiang in the same province?'', ``Pyotr Ilyich Tchaikovsky and Giuseppe Verdi have this profession in common''.  The information required to answer the question needs to come from two documents; specifically, the first document must provide information about the first entity/concept, while the second must provide information about the second entity/concept.

\paragraph{Temporal-evolution} A question that explores how something has changed, progressed, or developed over time. The first document covers the earlier historical period or initial state, while the second document covers the later historical period or final state.  For example: ``How has smartphone technology evolved over the past two decades?'', ``What changes have occurred in climate policy since the Paris Agreement?'', ``How has the portrayal of women in film changed from the 1950s to today?''. The answer should describe trends, shifts, or stages of development across a timeline.

\paragraph{Problem-solution} Questions that ask about both a problem and potential solutions. The first document details the problem and its implications, while the second document explores possible solutions or mitigation strategies. For example: ``What are the main challenges facing global food security, and what innovative agricultural technologies offer the most promising solutions?'', ``What are the causes and consequences of urban air pollution, and what policies have proven effective in reducing it?'', ``What factors contribute to the rise in mental health issues among teenagers, and what interventions can schools implement to support student well-being?'', ``Why is plastic waste a growing environmental concern, and what strategies can be used to reduce its impact?'', ``What are the limitations of current cybersecurity measures, and how can emerging technologies address them?''. 

\paragraph{Procedural} A question that asks how to do something or requests step-by-step instructions. For example: ``How do you train a neural network using TensorFlow?'', ``What are the steps to apply for a student visa to the United States?'', ``How can I set up a secure home Wi-Fi network?''. The answer should provide a clear, ordered list of steps or a detailed process to follow.

\paragraph{Causal} A question that seeks to understand why something happens or explores the relationship between cause and effect. For example: ``Why does increasing carbon dioxide in the atmosphere lead to global warming?'', ``What causes inflation to rise during economic booms?'', ``Why do some people develop allergies while others do not?''. The answer should explain the underlying reasons or mechanisms behind the phenomenon. 

\paragraph{Quantitative} A question that seeks numerical data, statistics, or measurements. For example: ``What is the average life expectancy in Japan?'', ``How many people use public transportation in New York City each day?'', ``What was the global GDP growth rate in 2023?''. The answer should include specific numbers, percentages, or quantitative comparisons supported by data.

\paragraph{Verification} A question that asks to confirm or deny the truth of a particular claim or statement. These questions involve evaluating one or more statements, where at least one is true and at least one is false. For example: ``Is it true that vitamin C cures the common cold and that antibiotics are effective against viruses?'', ``Did Einstein win the Nobel Prize for his theory of relativity and was he born in Austria?''. The answer should clearly indicate which statements are correct and which are incorrect, with justification.

\section{Fineweb-10BT Characteristic}
\label{appendix:fineweb_characteristic}
Fineweb-10BT is a 15-million-document web corpus that we embed with \textit{Snowflake Arctic-embed-m-v2.0} in the Pinecone vector database to use in our RAG pipeline. Using WebOrganizer’s \cite{weborganizer2024} 24‐label \textbf{TopicClassifier} and 24‐label \textbf{FormatClassifier}, we tagged every document along two orthogonal axes—\emph{topic} and \emph{document format}.  As the treemap in Figure \ref{fig:dist_of_topics_format} shows, the largest topical slices are \emph{Finance \& Business} (9.1\%), \emph{Sports \& Fitness} (7.6\%), and \emph{Entertainment} (7.6\%), while the dominant formats are \emph{News Article} (17.4\%), \emph{Personal Blog} (15.6\%), and \emph{Product Page} (14.9\%).  Crucially, long-tail domains (\emph{Software Dev.}, \emph{History}, \emph{Adult}, \emph{FAQ}, \emph{Legal}) remain present, giving the corpus the heterogeneity that retrieval-augmented models need for robust generalization.

\paragraph{\textbf{Internal benchmark.}} 
We constructed a diverse 500-question benchmark by sampling across the full 24$\times$24 topic–format grid, ensuring wide coverage of content types and styles.

\paragraph{\textbf{Query routing.}}
To improve retrieval speed, we map each topic to a dedicated Pinecone namespace. For each query, we use Falcon3-10B-Instruct with self-consistency voting to predict the relevant topics, querying only the top two namespaces.
\clearpage

\end{document}